\documentclass[letterpaper]{article}
\usepackage{uai2018}
\usepackage[margin=1in]{geometry}

\usepackage[T1]{fontenc}

\usepackage{algorithm}
\usepackage[noend]{algorithmic}
\algsetup{
    linenosize=\tiny,
    linenodelimiter=
}

\usepackage{amsmath,amssymb}
\usepackage{booktabs} 
\usepackage{multirow} 
\usepackage{graphicx}
\usepackage{url}
\usepackage{listings}
\usepackage{microtype}
\usepackage{subcaption}
\usepackage{tabularx}
\usepackage{xcolor}
\usepackage{xprintlen}

\usepackage{times}

\newcommand{\calX}{\mathcal{X}}
\newcommand*{\mydprime}{^{\prime\prime}\mkern-1.2mu}
\newcolumntype{Y}{>{\centering\arraybackslash}X}

\makeatletter
\newcommand\BeraMonottfamily{%
  \def\fvm@Scale{0.85}
  \fontfamily{fvm}\selectfont
}
\makeatother

\definecolor{anti-flashwhite}{rgb}{0.95, 0.95, 0.96}
\lstdefinestyle{code}{
  frame=tb,
  numbers=left,
  numberstyle=\tiny,
  xleftmargin=2.2em,
  framexleftmargin=2.2em,
  basicstyle=\footnotesize\BeraMonottfamily,
  showstringspaces=false,
  commentstyle=\color{gray}
}

\title{Fast Counting in Machine Learning Applications}

\author{
{\bf S. Karan}\\
University at Buffalo\\
skaran@buffalo.edu\\
\And
{\bf M. Eichhorn}\\
University at Buffalo\\
maeichho@buffalo.edu\\
\And
{\bf B. Hurlburt}\\
University at Buffalo\\
blakehur@buffalo.edu\\
\And
{\bf G. Iraci}\\
University at Buffalo\\
grantira@buffalo.edu\\
\And
{\bf J. Zola}\\
University at Buffalo\\
jzola@buffalo.edu\\
}

\begin{document}

\maketitle

\begin{abstract}
We propose scalable methods to execute counting queries in machine learning applications. To achieve memory and computational efficiency, we abstract counting queries and their context such that the counts can be aggregated as a stream. We demonstrate performance and scalability of the resulting approach on random queries, and through extensive experimentation using Bayesian networks learning and association rule mining. Our methods significantly outperform commonly used ADtrees and hash tables, and are practical alternatives for processing large-scale data.  
\end{abstract}

\section{INTRODUCTION}

Counting data records with instances that support some specific configuration of the selected variables is one of the basic operations utilized by machine learning (ML) algorithms. For example, when learning Bayesian network structure from data counting is necessary to evaluate a scoring function, or to assess constraints (e.g., via mutual information)~\cite{Koller2009}. In association rule mining, counting over binary data representing transactions is required to assess support and confidence for a given association rule~\cite{Agrawal1993}. Other examples include problems ranging from  classification~\cite{Kohavi1996,Quinlan1996} through deep learning~\cite{Lee2009,Salakhutdinov2009} to information retrieval~\cite{Ramos2003}.

While counting is typically viewed as a black-box procedure, and implemented using simple and not necessarily efficient strategies, e.g., contingency tables, in many practical applications it accounts for over 90\% of the total execution time (we show several practical cases in Sec.~\ref{sec:results}). Consequently, improving performance of counting can directly translate into better performance of these applications. At the same time, popular specialized approaches based on data indexes, such as ADtrees~\cite{Moore1998}, have limited applicability due to the significant preprocessing and memory overheads, which easily exceed the capability of current computational servers. This holds true for a broad spectrum of problem sizes and applications, with cases involving anywhere from tens to hundreds of variables, and thousands to millions of instances. As the size of the data analyzed by ML codes increases, there is a clear need for easy-to-adopt, efficient and scalable~counting~strategies.

In this paper, we address the above challenge by designing simple, yet fast and memory efficient counting strategies. Our methods are derived from the standard techniques like bitmap set representation and radix sorting, which can be efficiently implemented in a software. We describe an intuitive and convenient programming interface that leverages properties of the operators used in ML to separate the counting process from how the counts are utilized. This interface enables us to aggregate counts in a stream-like fashion. We encapsulate our methods in an open source software, and demonstrate its performance on random queries, Bayesian networks learning and association rule mining. Through extensive experiments on multiple popular benchmark data, we show that our strategies are orders of magnitude faster than the commonly used methods, such as ADtrees and~hash~tables.


\section{PRELIMINARIES}\label{sec:prelim}

\begin{figure*}[t]
\centering
\begin{subfigure}[b]{0.2\textwidth}
\centering
\includegraphics[scale=0.95]{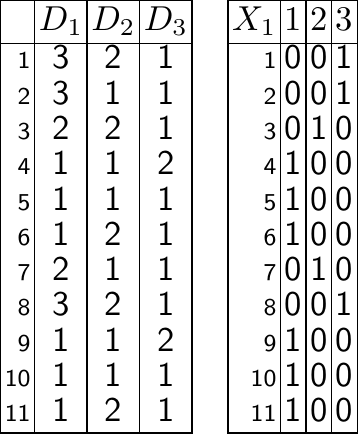}\caption{}
\end{subfigure}\hspace{10pt}
\begin{subfigure}[b]{0.4\textwidth}
\centering
\includegraphics[scale=0.575]{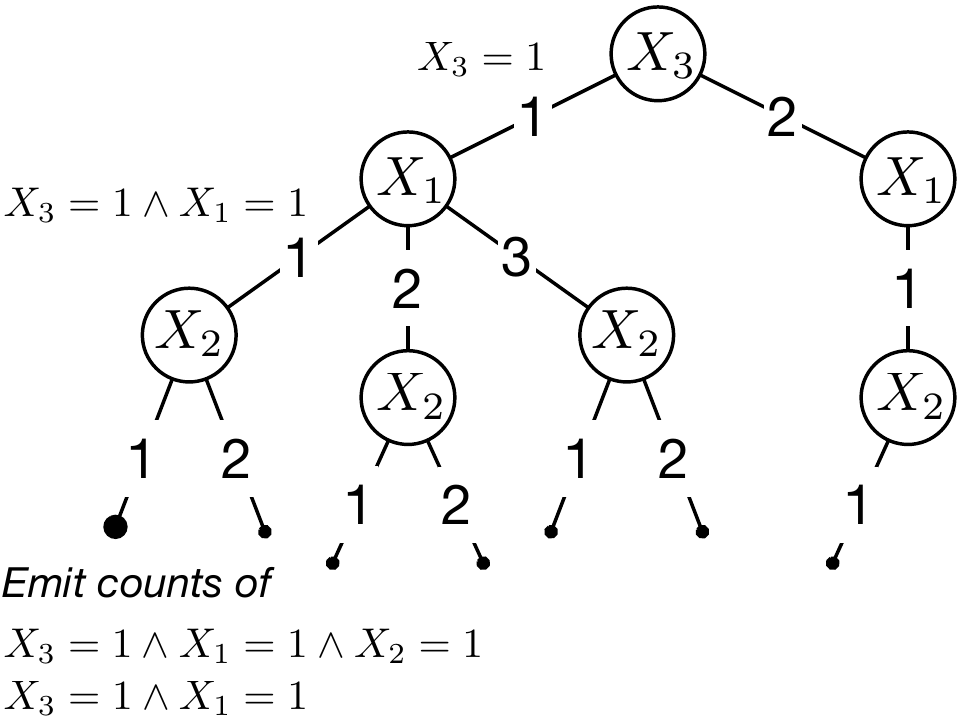}\caption{}
\end{subfigure}\hspace{10pt}
\begin{subfigure}[b]{0.3\textwidth}
\centering
\includegraphics[scale=0.575]{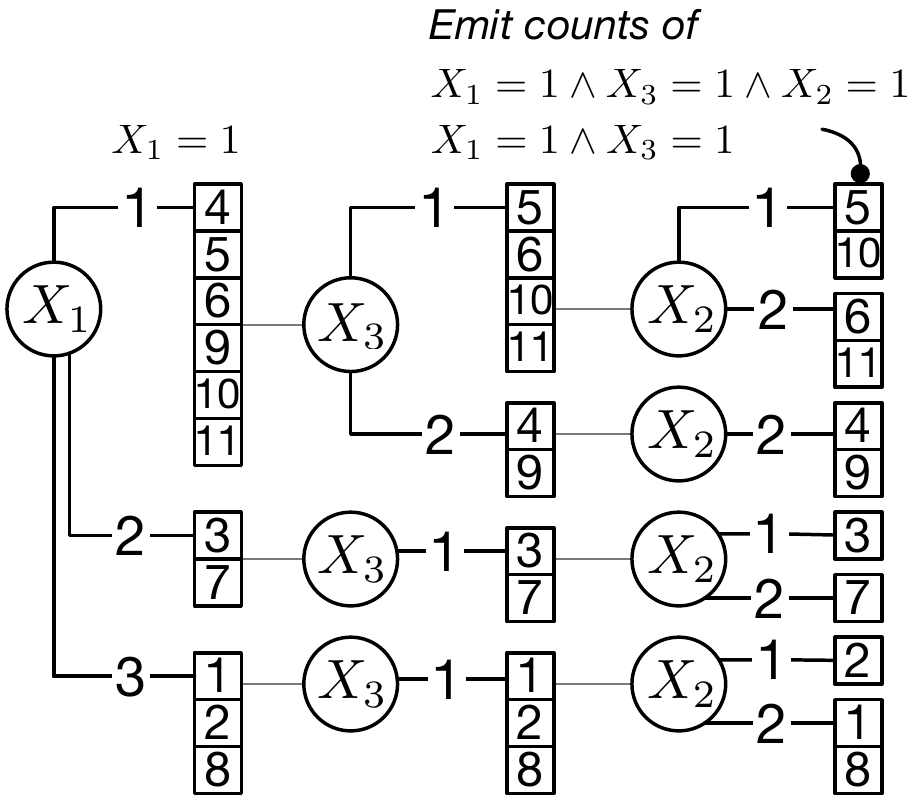}\caption{}
\end{subfigure}
\caption{(a) Database $D$ with three variables, and the corresponding bitmap representation of $X_1$. (b) Example of executing $Query(X_2,\{X_1,X_3\})$ over $D$ using Bitmap strategy, and (c) Radix strategy.}\label{fig:database}
\end{figure*}

Consider a set of $n$ categorical random variables $\calX = \{X_1, X_2, \dots, X_n\}$, where the domain of variable $X_i$ is represented by states $[x_{i1},\ldots,x_{ir_i}]$. Alternatively, we can think of $X_i$ as a symbolic feature with arity~$r_i$, and for convenience we can represent its states by integers $[1,\ldots,r_i]$. Let $D = [D_1, D_2, \ldots, D_n]$ be a complete database of instances of $\calX$, where $D_i$, $|D_i|=m$, records observed states~of~$X_i$. Given $D$, and a set of input variables $\{X_i, X_j, \ldots\} \subseteq \calX$, the counting query $Count((X_i=x_i) \land (X_j=x_j) \land \ldots)$ returns the size of the support in $D$ for the specific assignment $[x_i,x_j,\ldots]$ of variables $[X_i,X_j,\ldots]$. For example, for the database in Fig.~\ref{fig:database}, the query $Count((X_1=3) \land (X_2=2) \land (X_3=1))$ would return $2$, as there are $2$ instances matching the query condition. We note that the above formulation of counting is a special and simple case of the general counting problem in conjunctive queries, known from database theory~\cite{Greco2014} (we provide more details in Section~\ref{sec:related}).

Counting queries are the basic operations performed when learning statistical models from data. In some ML applications, such as association rule mining or learning probabilistic graphical models, they may account for over 90\% of the total execution time. In the most basic form, the queries can be issued without shared context. For example, to estimate joint probability $p(X_i=x_i,X_j=x_j)$ from $D$, we could use just one query: $\hat{p}(x_i,x_j|D) = \frac{Count((X_i=x_i) \land (X_j=x_j))}{m}$. However, in the most common use scenarios, a group of consecutive queries is executed over the same set of variables (i.e., the queries share context). For~instance, consider log-likelihood score frequently used in Bayesian networks learning~\cite{DeCampos2006}:
\begin{equation}\label{eq:likelihood}
\mathcal{L}(X_i|Pa(X_i))=\sum_{j=1}^{q_i}\sum_{k=1}^{r_i}N_{ijk}\log\left(\frac{N_{ijk}}{N_{ij}}\right),
\end{equation}
where $Pa(X_i) \subseteq \calX - \{X_i\}$ is a set of predictor variables for $X_i$, $j$~enumerates all possible $q_i = \prod_{X_j\in Pa(X_i)} r_j$ states of variables in $Pa(X_i)$, and $N_{ij}$ and $N_{ijk}$ are respectively the counts of instances in $D$ such that variables in $Pa(X_i)$ are in state $j$, and the counts of instances such that variables in $Pa(X_i)$ are in state $j$ and $X_i$ is in state~$k$. To compute $\mathcal{L}$ we require multiple counting queries over the same group of variables $Pa(X_i) \cup \{X_i\}$, testing different configurations of their states. Moreover, we care only about queries that return non-zero counts $N_{ijk}$ (note that non-zero $N_{ijk}$ implies non-zero $N_{ij}$), since only those contribute to the final sum.

The standard approach to handle queries that share context is to either directly scan the database $D$ to construct $r_i \times q_i$ contingency table of counts (or its high-dimensional variant such as data cube~\cite{Han2011}), or to first create an ADtree index to cache all sufficient statistics from $D$, and then to materialize contingency table on demand. Here materialization is done by retrieving the required counts via fast traversal over the index~\cite{Moore1998,Anderson1998}. However, both these approaches have significant limitations. 

To use a contingency table we have to either maintain a lookup table with $r_i \cdot q_i$ entries, or to use a dictionary (e.g., hash table) with keys over the states of $Pa(X_i)$ and values being vectors of counts for the corresponding states of $X_i$. While lookup table may offer very fast memory accesses during construction and querying phases, it becomes computationally impractical, since usually it is very sparse. This is because even for large $m$, most of the time $D$ will not contain all $q_i$ possible configurations for the majority of sets $Pa(X_i)$. Consequently, lookup tables become a feasible choice only when we are dealing with a small number of variables, each with very small arity. Dictionaries address the problem of sparsity, as they store only configurations that are observed in $D$. However, they impose non-trivial overheads owing to the cost of hashing in a hash table dictionary or traversing scattered memory in a search tree dictionary. Moreover, when large number of high-arity variables are considered, a dictionary quickly becomes memory intensive easily exceeding capacity of a typical \mbox{cache~memory.}

The alternative approach is to use one of many published variants of the ADtree index, e.g.,~\cite{Moore1998,Anderson1998,VanDam2008,VanDam2013}. Here the idea is to first invest (significant) time and memory to enumerate and cache counts of all configurations found in $D$, and then reference those counts to answer subsequent queries. However, even with various optimizations, the space complexity of ADtrees is exponential in the number of variables, and even for modestly sized $D$ it may exceed the available main memory. Moreover, by caching all counts indiscriminately, ADtrees often store entries that are never referenced in a given application, creating unnecessary memoization and searching overhead. Finally, ADtrees still require that a contingency table is materialized to deliver retrieved counts, and hence they pose a significant challenge in balancing memory and computations.

\section{PROPOSED APPROACH}

Given the database $D$, our goal is to provide memory and computationally efficient mechanism to answer counting queries with shared context. The memory efficiency is critical, since many ML algorithms, especially in classification and probabilistic graphical modeling, already have significant memory constraints (see for example~\cite{Yuan2011}). If the memory has to be devoted to handling queries instead of being used by the actual algorithm, it would clearly constrain the applicability of the algorithm. At this point it is worth noting that ML applications fall into a gray zone in terms of the size of the input data on which they typically operate. On the one hand, the size of the input is too small to benefit from many excellent optimizations known from database theory (some we review in Sec.~\ref{sec:related}), as those are targeting cases in which volume of the data necessitates concurrent use of both persistent and main memory. On the other hand, the data is too large to warrant efficient execution using direct techniques like simple contingency~tables.

To address this situation, we first define an intuitive programming interface to abstract the query context, including how counts are utilized by the target application. Then, we overlay the interface on top of two simple, yet very efficient, query execution~strategies, where instead of storing counts we {\it consume} them in a~stream-like~fashion.

\begin{figure}[ht]
\centering
\begin{lstlisting}[style=code, language=C++]
class L {
public:
  void operator()(int Nijk, int Nij) {
    double p = 1.0 * Nijk / Nij;
    score_ += (Nijk * log2(p));
  }
  double score() const { return score_; }
private:
  double score_ = 0.0;
};
\end{lstlisting}
\caption{Example C++ code packaging Eq.~(\ref{eq:likelihood}) into our programming interface.}\label{fig:code}
\end{figure}

\subsection{Programming Interface}\label{sec:interface}

In a typical application, counts provided by queries with shared context are iteratively aggregated via some associative and commutative operator. One simple example with the summation operator is given in Eq.~(\ref{eq:likelihood}). A more complex example could be Dirichlet priors with the product of gamma functions~\cite{Cooper1992}. From the computational point of view, this assumption is very helpful as it provides ample opportunities for optimization. We note also that while it may look very narrow, it actually accurately captures surprisingly many ML applications, which primarily involve estimating conditional probabilities. Examples include classifiers and regression, feature extraction, different variants of probabilistic graphical models, etc..

Following notation in Sec.~\ref{sec:prelim}, let us consider a set of variables $(Pa(X_i) \cup \{X_i\}) \subseteq \calX$ and their corresponding counts $N_{ij}$ and $N_{ijk}$, for some specific configuration $j$ of $Pa(X_i)$ and $k$ of $X_i$. Here we are distinguishing between counts for $(Pa(X_i) \cup \{X_i\})$ and $Pa(X_i)$ to simplify computing conditional probabilities while maintaining generality -- by passing $Pa(X_i) = \emptyset$ we can execute queries over single variable $X_i$, and by considering only $N_{ijk}$ we get joint queries $Pa(X_i) \cup \{X_i\}$. The key observation is that we can leverage associativity and commutativity, and instead of first gathering all counts and then performing aggregation, we can create a stream of counts corresponding to all unique and relevant configurations found in $D$, and perform the aggregation directly on the stream. This enables us to push computations to data, mitigating memory overheads due to counts caching. To achieve this, we abstract the computations via a function object (a concept supported by all modern programming languages), which is then repeatedly invoked over the stream. The example function object corresponding to Eq.~(\ref{eq:likelihood}) is given in Fig.~\ref{fig:code}. In the essence, the object receives $N_{ijk}$ and $N_{ij}$ via the function call operator (line 3), performs the required intermediate computations, and then aggregates the result into internal state. This internal state can be then inspected (line 7) to retrieve the final result of the aggregation. From the user perspective, the function call operator acts as an interface, and is directly invoked by a routine responsible for enumerating, and emitting, all unique configurations for the variables of interest (see parameter $F$ in Algs.~\ref{alg:bquery}~and~\ref{alg:radquery} in the following sections). Thus the interface provides a convenient encapsulation, and the end-user who defines the function object (e.g., implementing a scoring function in BN learning) can focus solely on expressing computations (i.e.,~high-level logic and correctness), and does not have to worry about potentially complex logic of low-level details (e.g., how counts are enumerated). Additionally, because function object behaves like a function, but has the advantage of possessing an internal state, it is a convenient mechanism to express even the most demanding~computations.

While the proposed interface stems from a relatively simple observation, it has several immediate advantages. First, by separating functionality (i.e., data traversing from computations) we gain flexibility to rapidly investigate different data traversal schemes to extract counts, or even alternate between different strategies depending on the query context (e.g., how many variables are involved, their domain, etc.). Second, since counts are aggregated into an isolated state represented by a function object, and multiple objects can coexist independently, multiple groups of queries, each group with individual context, can be executed concurrently and in parallel, e.g., by different threads. Collectively, this makes the proposed design extremely flexible, efficient and easy to use, as we demonstrate in the experimental results~section.

\subsection{Bitmap Strategy}\label{sec:BMap}

For the specific $X_i$ and $Pa(X_i)$ our task now is to enumerate counts $N_{ij}$ and $N_{ijk}$ for all configurations $j$ and $k$ found in $D$, and then pass the counts to a function object for aggregation. The idea behind the Bitmap strategy is to represent each variable $X_i$ via a set of $r_i$ bitmaps of size $m$, where each bitmap indicates instances for which $X_i$ is in the corresponding state (see Fig.~\ref{fig:database}a). Then, the entire process of enumerating counts can be reduced to performing logical \textsc{and} on bitmaps, equivalent of set intersection, and to bit counting, equivalent of computing set cardinality. This is summarized in Alg.~\ref{alg:bquery}, with example in Fig.~\ref{fig:database}b (for convenience, in the algorithm we use set notation instead of directly representing bitmaps).

\begin{algorithm}
  \small
  \caption{$\textsc{Query}(X_i, Pa, F, b)$}\label{alg:bquery}
  \begin{algorithmic}[1]
  \IF {$|Pa| = 0$}
    \STATE $N_{ij} \leftarrow |b|$
    \FOR {$v \in [1, \ldots, r_i]$}
       \STATE $b_v \leftarrow \{p\:|\,D_i[\,p\,] = v\}$
       \STATE $N_{ijk} \leftarrow |b \cap b_v|$
       \IF {$N_{ijk} > 0$}
         \STATE $F(N_{ijk}, N_{ij})$ \hspace{4pt}$\lhd$ \textsf{\scriptsize emit new configuration}
       \ENDIF
    \ENDFOR
   \ELSE
     \STATE $X_h \leftarrow \textsc{head}(Pa)$
     \FOR {$v \in [1, \ldots, r_h]$}
        \STATE $b_v \leftarrow \{p\:|\,D_h[\,p\,] = v\}$
        \IF {$|b \cap b_v| > 0$}
          \STATE $\textsc{Query}(X_i, \textsc{tail}(Pa), F, b \cap b_v)$
        \ENDIF
     \ENDFOR
  \ENDIF
  \end{algorithmic}
\end{algorithm}

To execute counting queries for $X_i$ and $Pa(X_i)$ (abbreviated to~$Pa$), and function object $F$, we perform Depth First Traversal (DFS) over the tree whose leaves represent all possible $r_i \cdot q_i$ states of interest (recall that $q_i = \prod_{X_j\in Pa(X_i)} r_j$). The bottom layer of the tree is induced by the states of~$X_i$, and the top layers correspond to variables in $Pa$. When moving down the tree (lines 9-13), we compute intersection between the set of instances supporting variables' configurations seen thus far (in the algorithm denoted by $b$, which initially consists of all $m$ instances), and the set of instances supporting current configuration of the considered variable from $Pa$ (in the algorithm denoted by~$b_v$). We continue traversal only if the size of the intersection is greater than zero, implying non-zero count for given joint configuration of variables. Once we reach a leaf of the tree (lines 1-7), we compute the final counts $N_{ijk}$ and $N_{ij}$ for the corresponding configurations $j$ and $k$, and emit those via~call~to~$F$. 

The depth of the tree depends on the number of variables involved in the query, and the number of leaves is bounded by $O(\min(q_i,m))$, with each step in the traversal involving $O(m)$ cost of computing intersection and cardinality. While the tree could potentially involve exponential (in~the~number of query variables) number of nodes, it is never explicitly stored in the memory, and even for $D$ with large number of instances many configurations have zero count, allowing for their corresponding sub-trees to be pruned. To further leverage this property, we order $Pa$ such that variables with lowest entropy estimated from $D$ are at the top of the tree. Since variables with low entropy are likely to have configurations for which there will be only few supporting instances, they are more likely to trigger zero counts and hence lead to a smaller tree to traverse. For example, consider executing $Query(X_2, \{X_1,X_3\})$ outlined in Fig.~\ref{fig:database}b. There are total of $7$ configurations which we should enumerate, and if we traverse the tree starting from variable $X_3$, which has lower entropy than $X_1$, then we will have to consider $6$ intermediate states. If we were to start with variable $X_1$, then this number would increase to $9$. This optimization performs extremely well in practice, and, as we show in the experimental results section, for certain ranges of $n$ and $m$, Bitmap outperforms other strategies.

In the practical terms, the strategy can be efficiently implemented using streaming extensions (SIMD) in current processors. Bitmaps for individual variables can be precomputed and laid out in the memory instead of $D$, with acceptable memory overhead (i.e., $m \cdot r_i$ vs. $m \cdot \log_2(r_i)$ bits), and the relative ordering of variables in $D$, based on their entropy, can be established beforehand as well.

\subsection{Radix Strategy}\label{sec:rad}

While the Bitmap strategy is amenable to very efficient implementation, its scalability may still suffer when datasets with very large number of instances are exercised by queries with many variables, or variables with high arity. This is because in such cases the DFS tree will have fewer nodes to prune, and the advantage of fast bit-wise operations will be offset by the poor asymptotic behavior. To address these cases, we consider an alternative approach, which we refer to as Radix strategy. The strategy is derived from the classic radix sort algorithm, and it involves recursively partitioning instances in $D$ such that single partition at given level captures all instances corresponding to one specific configuration of the query variables. This approach is summarized in Algs.~\ref{alg:radquery}~and~\ref{alg:radix}, with example~in~Fig.~\ref{fig:database}c.

\begin{algorithm}
  \small
  \caption{$\textsc{Query}(X_i, Pa, F)$}\label{alg:radquery}		
  \begin{algorithmic}[1]
    \STATE {$B \leftarrow [[1, \ldots, m]]$}
    
    \IF {$|Pa| \neq 0$}
        \STATE $B \leftarrow \textsc{Buckets}(\textsc{head}(Pa), \textsc{tail}(Pa), \textsc{head}(B))$
    \ENDIF
    
    \FOR {$b \in B$}
   		\STATE {$N_{ij} \leftarrow |b|$}
        \IF {$N_{ij} >0$}
    		\STATE $B^\prime \leftarrow \textsc{Buckets}(X_i, [], b)$
			\FOR {$b^\prime \in B^\prime$}
    			\STATE {$N_{ijk} \leftarrow |b^\prime|$}
            	\IF {$N_{ijk} > 0$}
                 	\STATE $F(N_{ijk}, N_{ij})$ \hspace{4pt}$\lhd$ \textsf{\scriptsize emit new configuration}
        		\ENDIF
          	\ENDFOR
        \ENDIF
  	\ENDFOR
 	\end{algorithmic}
\end{algorithm}

\begin{algorithm}
  \small
  \caption{\textsc{Buckets($X_p$, $Pa$, $b$)}}\label{alg:radix}
  \begin{algorithmic}[1]
  	\STATE $B^\prime \leftarrow []$
    
   	\FOR {$q \in [1, \ldots, |b|]$}
    	\STATE $x_p \leftarrow D_p[b[q]]$
        \STATE $B^\prime[x_p].\textsc{append}(b[q])$
  	\ENDFOR
    
    \IF {$Pa = []$}
    	\RETURN $B^\prime$
   	\ENDIF
    
    \STATE $B\mydprime \leftarrow []$
    
    \FOR {$b^\prime \in B^\prime$}
    	\STATE $B\mydprime.\textsc{append}(\textsc{Buckets}(\textsc{head}(Pa), \textsc{tail}(Pa), b^\prime))$
   	\ENDFOR
    
    \RETURN $B\mydprime$
  \end{algorithmic}
\end{algorithm}

The algorithm follows the Most Significant Digit (MSD) radix, with the left most digits being states of individual variables in~$Pa$, and the least significant digit representing states of~$X_i$ (Alg.~\ref{alg:radquery}, line~3). At each level, the number of newly created partitions is proportional to the arity of the considered variable, and the size of the partition is the support in $D$ for the particular configuration. The order in which variables from $Pa$ are processed is not significant, since the cost of identifying empty partitions does not induce overheads. Because the actual instances in $D$ are not to be sorted, but only partitioned, it is sufficient that we maintain a list (in algorithms denoted by $B$) of partition descriptors containing indexes of the constituent instances and partition size (Alg.~\ref{alg:radix}, lines 1-4). As soon as all partitions prescribed by $Pa$ are identified we can proceed to emitting counts (Alg.~\ref{alg:radquery}, lines 4-11), which must be preceded by the final round of partitioning with respect to $X_i$ (Alg.~\ref{alg:radquery}, line 7).

The algorithm requires that for each variable $X_p \in Pa$ its corresponding data vector $D_p$ is completely scanned, leading to the overall $O(|Pa|\cdot m)$ complexity. In practice, the entire method is efficiently implemented by first organizing the database $D$ in the column-major format, and then maintaining a FIFO queue of partition descriptors, with $O(m)$ auxiliary space to keep track of the assignment of indexes to partitions. Moreover, partitioning for individual variables can be precomputed in advance, further bootstrapping the first step of the~algorithm.

To conclude the presentation, we note that both Bitmap and Radix strategies can be further augmented such that instead of enumerating all counts (i.e., executing queries with shared context) they deliver counts just for the specific assignment of the query variables. To achieve this, it is sufficient to process only a single path from the root to the leaf with the target assignment in the DFS tree of the Bitmap strategy, and to find the partition corresponding to the assignment, instead of all partitions, in~Radix.

\section{EXPERIMENTAL VALIDATION}\label{sec:results}

We implemented both proposed strategies as a C++ software library, which we complemented with Python bindings for the ease of use. At its core, the library uses standard SSE SIMD intrinsics to implement basic bitmap operations (i.e., logical \textsc{and}, and bit counting), and it exposes all functionality via the interface described in Sec.~\ref{sec:interface}. The resulting open source package, which we call
\texttt{\small SABNAtk},
is available from: 
\url{https://gitlab.com/SCoRe-Group/SABNAtk}.

We deployed \texttt{\small SABNAtk}
on a server with two Intel Xeon E5-2650 2.30~GHz 10-core CPUs, and 64~GB of RAM. To test the performance, we ran a series of experiments using popular ML benchmark datasets (see data summary in Tab.~\ref{tab:data}). For reference, we used hash table from the C++ standard library, and the sparse ADtree data index~\cite{Moore1998}. The hash table represented contingency table created by directly scanning the input database, with keys encoding specific assignment of variables in $Pa(X_i)$, and values representing vectors of counts for specific assignment of $X_i$. To maximize cache memory usage, the strategy operated on the database stored in the row-major order. Finally, to make the comparison fair and avoid biases due to the differences in programming languages, we developed an efficient ADtree implementation in C++. We note that other available implementations, for example~\cite{ADTPY}, turned out to be substantially slower than our version.

\begin{table}
  \caption{Benchmark data used in experiments.}\label{tab:data}
{\scriptsize\sf
\begin{tabularx}{\columnwidth}{Xlll}
	\toprule
	Dataset     	&   $n$ 	&  Range of $r_i$ 	& Average $r_i$ \\
	\midrule
	Child    		&   20  	&   2-6				& 3	\\
	Insur(ance)   	&   27  	&   2-5       		& 3.3 \\	
	Mild(ew)   		&   35  	&   3-99       		& 16.4\\
	Alarm       	&   37  	&   2-4        		& 2.83 \\
    Barley       	&   46  	&   3-67      	 	& 9.02 \\
	Hail(finder)	&   56  	&   2-11      		& 3.98 \\
    Win95(pts)		&   74  	&   2-2      		& 2 \\
	Path(finder)	&   104  	&   2-63      		& 4.2 \\
\bottomrule
\end{tabularx}
}
\end{table}
\begin{figure}
  \scriptsize\sf
  \centering
  \setlength{\tabcolsep}{2pt}
  \setlength\extrarowheight{4pt}
  \renewcommand\tabularxcolumn[1]{m{#1}}
  \hspace*{-10pt}
  \begin{tabularx}{\columnwidth}{r|XXX}
  $m$ & Child $n=20$ & Alarm $n=37$ & Hail $n=56$ \\
  \hline
  1K & \includegraphics{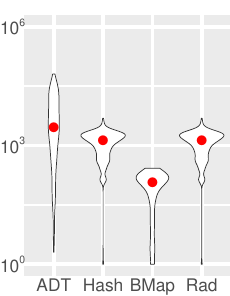} & \includegraphics{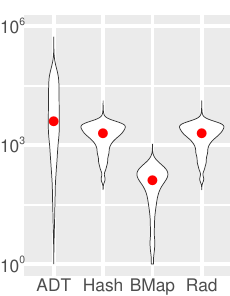} & \includegraphics{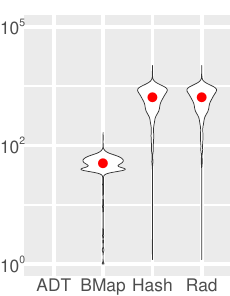} \\
  10K & \includegraphics{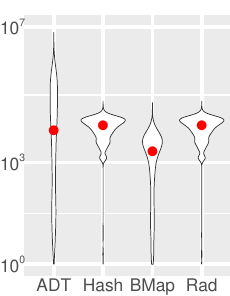} & \includegraphics{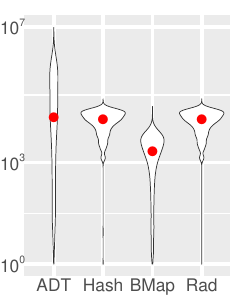} & \includegraphics{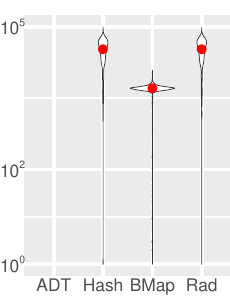} \\
  100K & \includegraphics{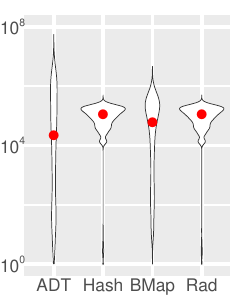} & \includegraphics{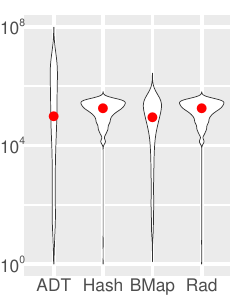} & \includegraphics{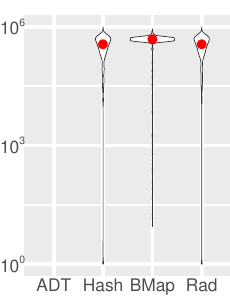} \\
  1M & \includegraphics{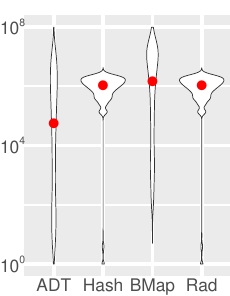} & \includegraphics{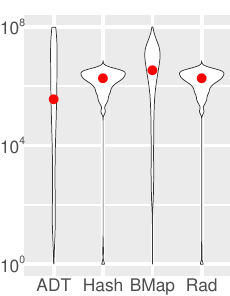} & \includegraphics{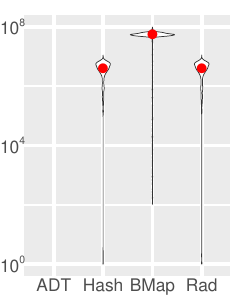} \\
  10M & \includegraphics{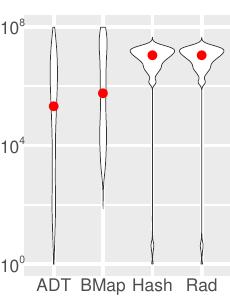} & \includegraphics{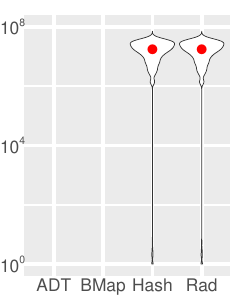} & \includegraphics{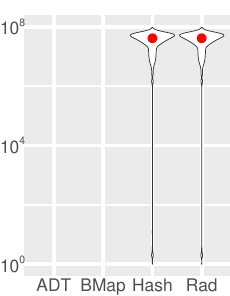} \\
  \end{tabularx}
  \caption{Comparison ADT, Hash, BMap and Rad strategies on the stream of uniformly random queries. The plots show the distribution of response time in microseconds, computed from the same sample of 1,000 queries for different number of instances~$m$. \mbox{Y-axis} is in log$_{10}$-scale.}\label{fig:resrandom1}
\end{figure}

In the following, we discuss in detail several key results obtained using the above setup. More extensive results (including additional test cases), together with the data that can be used to reproduce our experiments, are available from: {\small \url{https://gitlab.com/SCoRe-Group/SABNAtk-Benchmarks}}.

Before we proceed with the results discussion, we note that in order to use ADtree, the input database has to be first indexed. In all our experiments, we considered only the query time with index already loaded into memory. Moreover, ADtree provides a hyper-parameter $\ell$ to configure the size of the leaf-lists~\cite{Moore1998}. We experimented with several values of the parameter, to fine-tune the trade-off between query performance and memory consumption, and we settled with $\ell=16$, which we use throughout the paper. The results obtained for other ADtree configurations exhibited similar patterns to those reported in the paper, and are available online.

\subsection{Random Queries}

\begin{figure}
  \centering
	\includegraphics{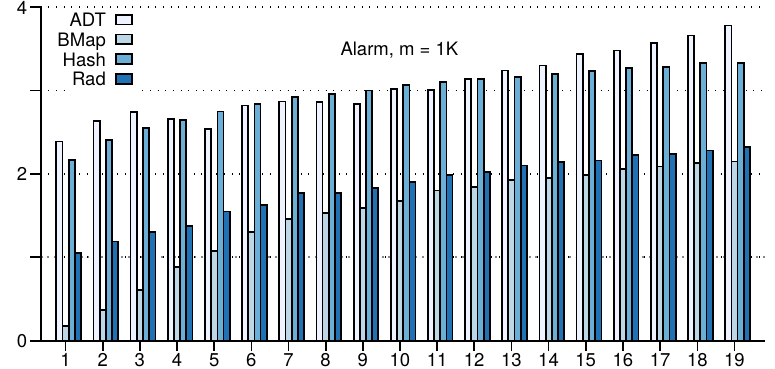} \\
    ~\\
   	\includegraphics{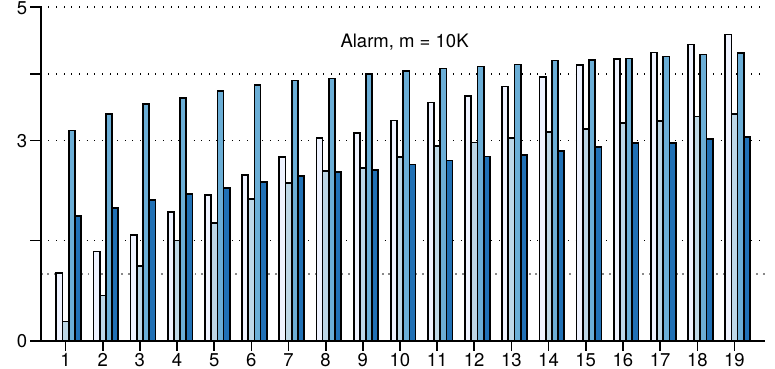} \\
    ~\\
   	\includegraphics{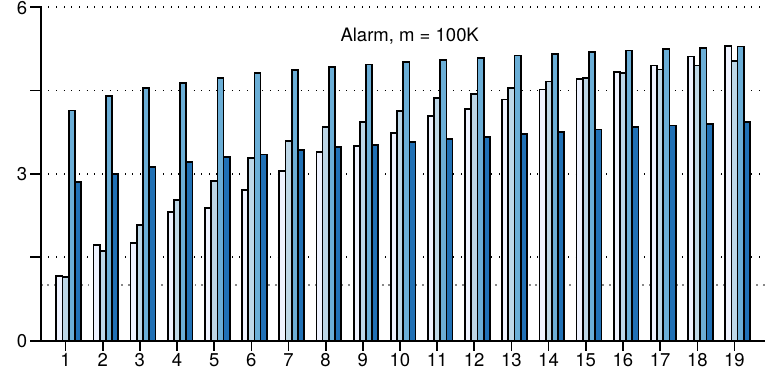} \\
    ~\\
    \includegraphics{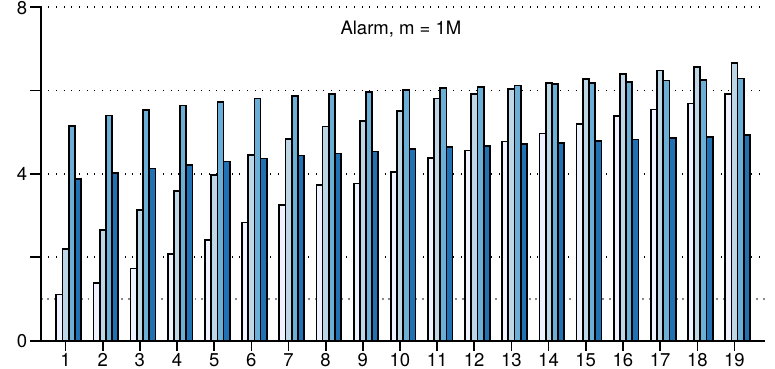} \\
    ~\\
    \includegraphics{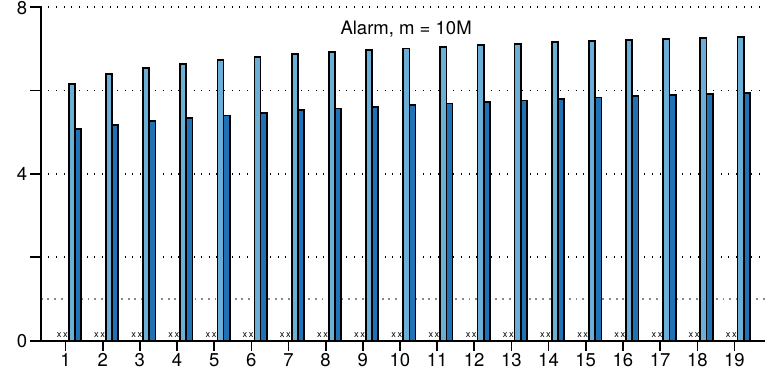} \\
  \caption{Comparison of ADT, Hash, BMap, and Rad for different sizes of $Pa$ (x-axis) in the sample of 1,000 uniformly random queries. The plot shows the average query response time in microseconds. Y-axis is in \mbox{log$_{10}$-scale}.}\label{fig:resrandom2}
\end{figure}

In the first set of experiments we tested how ADtree (ADT), HashTable (Hash), Bitmap (BMap), and Radix (Rad) strategies respond to a stream of random queries. The idea here is to understand average performance of each strategy in case where we have no prior information about specific query execution patterns. For each benchmark database, we generated 100,000 queries of the form $Query(X_i,Pa(X_i))$ as follows. First, the size of $Pa(X_i)$ was sampled uniformly from the range $[1,\ldots, n-1]$, and then variables were assigned to $X_i$ and $Pa(X_i)$ by randomly sampling without replacement from $\calX$. To measure time taken to execute the query, we used a simple function object that consumes and immediately discards the counts. In this way, the overhead of performing computations on the counts was negligible, and did not offset the actual time spent by each strategy to enumerate the counts. Each query was executed five times to obtain the average response time (with negligible variance), and exactly the same stream of queries was processed by each strategy. The results of this experiment are summarized in~Figs.~\ref{fig:resrandom1}~and~\ref{fig:resrandom2}. Here, we note that plots are in $\log_{10}$ scale, and should be interpreted with care.

Figure~\ref{fig:resrandom1} shows that depending on the number of input variables $n$, and the number of instances $m$, different strategies perform better in terms of the mean response time. When the number of instances is relatively small, Bitmap strategy significantly outperforms other methods. This is explained by two factors: first, for small scale data, Bitmap benefits from continuous memory accesses, and acceleration via SIMD instructions, second, because in small datasets many possible variable configurations are unobserved, Bitmap is able to prune significant portions of the DFS tree, taking advantage of the entropy-based data reordering (see Sec.~\ref{sec:BMap}). However, as the number of instances in the input database increases these advantages diminish, to the extent where the average time taken by a query becomes unacceptable (longer than several seconds, a threshold we set to make experiments computationally feasible). The Radix and HashTable strategies perform steadily across all datasets, and are able to handle even the most demanding test cases. This is expected, since both strategies involve similar data access pattern (i.e., scanning selected columns of the input database). However, Rad is on average 20 times faster than Hash (not captured in the figure due to log-scale), as it does not require costly hashing and scattered memory accesses.

The ADtree strategy exhibits the best mean response for problems with few variables and large number of instances, but it significantly underperforms in all remaining test cases. In fact, as the number of variables increases, ADtree fails to index the database and cannot be used to answer the queries. This is because the exponential growth of the number of configurations, which have to be cached, leads to the exhaustion of the main memory. Recall also that we do not include ADtree preprocessing time, which for datasets with more than 100K instances exceeds several hours, much longer than the time required to answer all 100K queries.



To further dissect performance of random queries, in Fig.~\ref{fig:resrandom2} we show how the response time varies with the number of query variables, for an example database. When processing small queries ($|Pa| < 4$), ADtree is generally outperformed by BMap, and when handling large queries ($|Pa| > 10$) it is slower than Rad. Moreover, the cost of Radix strategy is linear with the query size, compared to the exponential growth of ADtree.

Based on the tests with random queries, we conclude that Bitmap and Radix strategies significantly outperform ADtree and HashTable, except of a small set of scenarios in which small queries are executed over databases with few variables and millions of instances, if we exclude the preprocessing time.

\subsection{Queries in Bayesian Networks Learning}

Counting queries with shared context are the key operations performed in score-based Bayesian networks structure learning and Markov blankets discovery~\cite{Koller2009}. Both problems depend on the parent set assignment as a sub-routine~\cite{Koivisto2006}, and for given $X_i$ can be solved exactly by traversing a lattice with $n$ levels formed by the partial order \textit{set inclusion} on the power set of $\calX - \{X_i\}$. For given $\calX$ and $D$, queries of the form $Query(X_i, Pa, F)$ are performed for each $X_i$, where $Pa$ iterates over all possible subsets of $\calX - \{X_i\}$, starting from empty set. Hence, at level $i=0,\ldots,n-1$ we have that $|Pa|=i$, and there are total $\binom{n - 1}{i}$ queries to execute, creating interesting pattern of queries that grow in size as computations progress. The function object $F$ implements decomposable scoring function, e.g., MDL~\cite{Schwarz1978}, BDeu~\cite{Cooper1992}, etc., that evaluates the~assignment~of~$Pa$~as~parents~of~$X_i$.

We used all tested strategies to implement counting queries in the open source parent set assignment solver~\cite{SABNA}. The solver uses MDL scoring function, deploys several optimizations to eliminate some of the queries based on the results seen thus far, and because it effectively explores large combinatorial search space it has significant memory requirements. It also leverages OpenMP to execute multiple queries in parallel. As such, it serves as a practical benchmark for the query strategies. In our experiments, instead of considering all possible parent set sizes, as required by the exact solver, we limited the solver to $|Pa| \leq 6$, to make tests computationally feasible. This corresponds to a heuristic in which we make an assumption that no variable in the final Bayesian network can have more than six parents.

\begin{figure}
  \centering
  \includegraphics{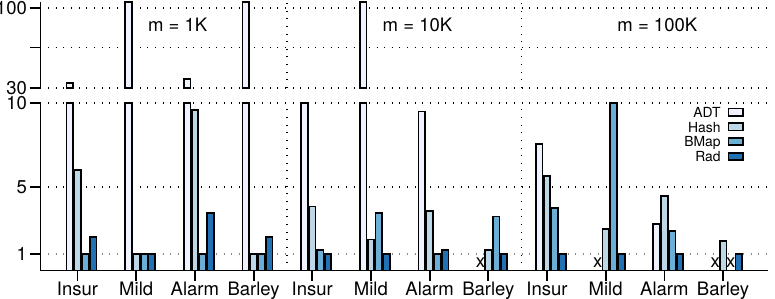}
  \caption{The total execution time of the parent set assignment solver with ADT, Hash, BMap and Rad strategies, normalized with respect to the fastest method. The solver was executed up to the level where $|Pa|=6$.}\label{fig:mpsbuild1}
\end{figure}

Figure~\ref{fig:mpsbuild1} shows the total execution time of the solver for different input databases and query strategies. From the plots, we can see that our proposed strategies significantly outperform ADtree and HashTable, across all benchmarks. In fact, for datasets with high-arity variables, i.e., Mildew and Barley, the Radix strategy is 100 times faster than~ADtree. This is explained by very large number of states that are to be expected in such datasets (and are costly to manage by ADtree), and by the pattern of how queries are generated by the solver. Because the size of the queries and their number grow together, there are only a few small queries that benefit ADtree, and increasing number of queries that are easily handled by the~Radix~strategy.

To illustrate how critical is the performance of counting queries for parent sets assignment, in Tab.~\ref{tab:data_time} we report the total execution time of the solver, together with the fraction of the time taken by the queries, when running on databases with 100K instances. In all cases, the execution is dominated by database querying that accounts for 90\%-99\% of the total time. Interestingly, this fraction is smaller for ADtree than for other strategies, even though ADtree is slower (we observed this pattern in all test cases). We~believe~that this is because BMap and Rad are memory friendly, and have minimal effect on memory utilization by the solver, thus minimizing cache update overheads, which in turn could slow down the solver. This is not the case for ADtree, which requires gigabytes of memory to run, and hence influences performance of the solver, affecting the ratio between the query and the~solver~time.

\begin{table}
  \caption{Execution time of the parent set assignment.}\label{tab:data_time}
  {\scriptsize\sf
  \begin{tabularx}{\columnwidth}{lXXXX}
	\toprule
		 	&   Insur &   Mild & Alarm & Barley\\
	\midrule
   
    	 \multirow{2}{*}{ADT}	& 35m20s    	&   \multirow{2}{*}{--}    	&   119m			& 	\multirow{2}{*}{--} \\
         						& $90.2\%$    	&   	 	  				&   $98.2\%$		& 		 				\\
         \hline
         \multirow{2}{*}{Hash}	& 26m26s    	&   		92m3s  	  		&   190m47s			& 		276m16s 		\\
         						& $98.2\%$    	&   		$99.1\%$  		&   $99.4\%$		& 		$98.2\%$ 		\\
        \hline
    	 \multirow{2}{*}{BMap}	& 17m28s    	&   		1107m57s  		&   100m55s			& 	\multirow{2}{*}{--}	\\
         						& $98.8\%$    	&   		$99.8\%$  		&   $99.9\%$		& 						\\
       	 \hline
         \multirow{2}{*}{Rad}	& 4m41s    		&   		37m28s    		&   43m4s			& 		156m32s 		\\
         						& $99.9\%$    	&   		$99.9\%$  		&   $99.9\%$		& 		$99.9\%$ 		\\
\bottomrule
\end{tabularx}
}
\end{table}

\subsection{Queries in Association Rule Mining}

Association rule mining is the classic method for establishing implication rules between a set of items in a database~\cite{Agrawal1993,Zaki2000}. Given a set of binary variables $\calX$, and a database of transactions $D$, where $D_i$ shows in which transactions item represented by $X_i$ was involved (i.e., $X_i$ is in state $1$), we want to identify rules of the form $Pa(X_i) \Rightarrow X_i$ with support, i.e., how frequently $Pa(X_i) \cup \{X_i\}$ are set together in~$D$, and confidence, i.e., how frequently the rule is true in~$D$, above some predefined thresholds. In the most direct form, the problem can be solved by traversing the power set lattice over $\calX$, a query pattern similar to the one used by the parent set assignment solver. However, compared to the parent set assignment, the actual queries are simpler, since we only require the counts of query variables being in state $1$. For example, to assess the rule $\{X_1, X_2\} \Rightarrow X_3$ we would perform queries $Count(X_1=1,X_2=1,X_3=1)$ and $Count(X_1=1,X_2=1)$. Consequently, instead of considering the query context, it is sufficient that our strategies search for one specific configuration of the query variables (as explained at the end of Sec.~\ref{sec:rad}).

\begin{figure}
  \centering
	\includegraphics{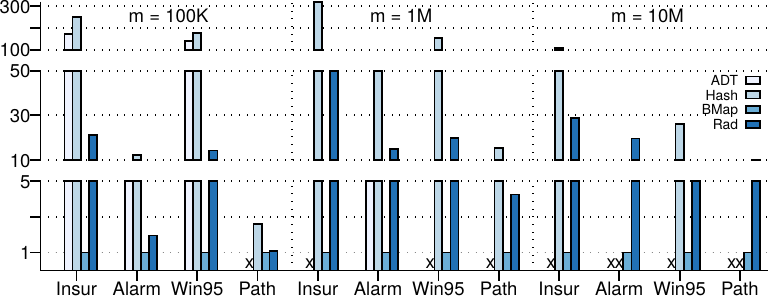}
  \caption{The total execution time of association rule mining with ADT, Hash, BMap and Rad strategies, normalized with respect to the fastest method. The solver was executed up to the level where $|Pa|=6$.}\label{fig:rules}
\end{figure}

We used all four tested strategies to implement simple association rule mining engine based on the bottom-up search~\cite{Zaki2000}. With the engine, we processed several large databases to enumerate rules with the support above $0.2$ and confidence above $0.3$, but of size less than seven. We selected these thresholds empirically to retrieve association rules with more than four variables, which allowed us to reliably measure the execution times (for smaller rules, the solver ran extremely fast). Results of this experiment are summarized in~Fig.~\ref{fig:rules}.

The figure shows that the Bitmap strategy significantly outperforms other approaches across all tested databases. Since in the Bitmap, processing queries with specific assignment amounts to a series of $|Pa(X_i)| + 1$ bitmap intersections, followed by bit counting, the cost of queries becomes linear in the size of the query and the number of instances in the database. Moreover, because all bit-wise operations are implemented via SIMD extensions, BMap becomes much faster than Rad, which has the same asymptotic behavior but involves less cache friendly operations. Finally, the overheads of traversing ADtree and handling its MCV subtrees (see~\cite{Anderson1998} for details) leads to its poor overall performance. At this point we should note that originally ADtree was designed for learning of association rules, however the design did not account for the memory and SIMD capabilities of modern processors.

\section{RELATED WORK}\label{sec:related}

As we mentioned through out the paper, counting queries in machine learning applications are often handled via some variant of the ADtree data index. The sparse ADtree~\cite{Moore1998,Anderson1998}, which we used in our experiments, precomputes and caches counts for all possible variable configurations. The counts are organized into a tree of \textit{vary nodes}, encoding the choice of variables to facilitate fast searching, and \textit{AD nodes} that store the actual query counts. To partially mitigate the excessive memory use, ADtrees do not explicitly represent most commonly occurring counts, and instead of creating AD nodes for counts lower than certain threshold, they resort to on-demand counting when such nodes are accessed. These base ideas have been extended by multiple researchers to account for dynamic data (i.e., updates to the database)~\cite{Roure2006}, and to improve performance on high-arity data~\cite{VanDam2008,VanDam2013}. However, as the core functionality in these data structures remains exactly the same, they suffer from the same limitations that we demonstrated in our experiments (expensive preprocessing, large memory footprint, significant traversing~overheads). 

Support for counting queries is a primary component in any database management system. In such systems, the query mechanism must support conjunctive queries over multiple tables, and with a variety of possible query predicates~\cite{Greco2014}. Moreover, the queries are typically executed over tables that cannot be fully materialized in the main memory. Our Bitmap strategy can be viewed as a practical realization of the Leapfrog Trie Join~\cite{Veldhuizen2012} with an unary relation, under assumption that the entire database resides in the main memory. 

The idea of using bitmaps to represent sets and their operations (e.g., intersection, cardinality, etc.) is frequent in software and databases design. This is because it allows to reduce memory, storage or network bandwidth, while maintaining the basic sets functionality~\cite{Kaser2016}. In these applications, bitmaps are typically compressed following methods like for example RLE encoding or Roaring~\cite{Lemire2010,Lemire2018}. The compressed bitmaps are orthogonal to our approach, and in fact we could use them to improve memory profile of our Bitmap strategy. However, as the compression induces computational overheads, and the size of the databases we consider practical is relatively small, currently we do not use compression.




\section{CONCLUSIONS}

In this paper, we describe efficient strategies for handling counting queries in machine learning applications. By combining convenient programming interface with memory efficient data traversing algorithms we are able to scale to large data instances, which we confirm via extensive experiments. The proposed solutions outperform and can substitute popular ADtree index. Moreover, to maintain best possible performance across different data instances, they can be selectively applied at the runtime depending on the properties of the queries.

While our approach is presented as a method for static databases, we note that it can be easily adopted to the cases where the input database expands with new instances during processing. This would amount to a simple update to the bitmaps in the Bitmap strategy, and is automatically handled in the Radix strategy.

\subsubsection*{Acknowledgements}

Authors wish to acknowledge hardware and technical support provided by the Center for Computational Research at the University at Buffalo. This work has been supported in part by the Department of Veterans Affairs under grant 7D0084, and by the National Institutes of Health under grant 1UL1 TR001412-01.


\bibliographystyle{IEEEtran}
\bibliography{references}

\end{document}